\documentclass[letterpaper, 10 pt, journal, twoside]{IEEEtran}

\IEEEoverridecommandlockouts                              

\usepackage{times}
\usepackage{epsfig}
\usepackage{graphicx}
\usepackage{amsmath}
\usepackage{amssymb}
\usepackage[noend]{algpseudocode}
\usepackage{balance}
\usepackage{xcolor}
\usepackage{url}
\usepackage{authblk}
\usepackage{multirow}
\newcommand{\eg}{\emph{e.g.},}
\newcommand{\ie}{\emph{i.e.},}
\newcommand{\etal}{\emph{et~al.}}

\usepackage{tabularx}
\usepackage{cite}
\usepackage[linesnumbered,ruled]{algorithm2e}
\usepackage[flushleft]{threeparttable}

\makeatletter
\def\BState{\State\hskip-\ALG@thistlm}
\makeatother
\algdef{SE}[DOWHILE]{Do}{doWhile}{\algorithmicdo}[1]{\algorithmicwhile\ #1}

\graphicspath{ {figures/} }

\title{\LARGE \bf
{Spatiotemporal Camera-LiDAR Calibration: \\A Targetless and Structureless Approach}
}

\author{Chanoh Park$^{1,2}$, Peyman Moghadam$^{1,2}$, Soohwan Kim$^{3}$, Sridha Sridharan$^{2}$, Clinton Fookes$^{2}$   
\thanks{Manuscript received: September, 10, 2019; Revised December, 2, 2019; Accepted January, 8, 2020.}
\thanks{This paper was recommended for publication by Editor Eric Marchand upon evaluation of the Associate Editor and Reviewers' comments.}
\thanks{
$^1$ Chanoh Park, Peyman Moghadam are with the Robotics and Autonomous Systems, DATA61, CSIRO, Brisbane, QLD 4069, Australia and the School of Electrical Engineering and Robotics, Queensland University of Technology (QUT), Brisbane, Australia.
E-mails: {\tt\footnotesize \emph{Chanoh.Park, Peyman.Moghadam}@data61.csiro.au} }
\thanks{
$^{2}$ Sridha Sridharan, Clinton Fookes are with the School of Electrical Engineering and Robotics, Queensland University of Technology (QUT), Brisbane, Australia.
E-mails: {\tt\footnotesize \emph{chanoh.park, peyman.moghadam, s.sridharan, c.fookes}@qut.edu.au}}
\thanks{
$^{3}$ Soohwan Kim is with Division of Smart Automotive Engineering, Sun Moon University, South Korea, E-mail: {\tt\footnotesize \emph{kimsoohwan}@gmail.com}}
\thanks{Digital Object Identifier (DOI): see top of this page.}
}

\begin{document}

\maketitle

\begin{abstract}
The demand for multimodal sensing systems for robotics is growing due to the increase in robustness, reliability and accuracy offered by these systems. These systems also need to be spatially and temporally co-registered to be effective. 
In this paper, we propose a targetless and structureless spatiotemporal camera-LiDAR calibration method.
Our method combines a closed-form solution with a modified structureless bundle adjustment where the coarse-to-fine approach does not {require} an initial guess on the spatiotemporal parameters. Also, as 3D features (structure) are calculated from triangulation only, there is no need to have a calibration target or to match 2D features with the 3D point cloud which provides flexibility in the calibration process and sensor configuration.
We demonstrate the accuracy and robustness of the proposed method through both simulation and real data experiments using multiple sensor payload configurations mounted to hand-held, aerial and legged robot systems. Also, qualitative results are given in the form of a colorized point cloud visualization.  
 
\end{abstract}


\section{Introduction}
\label{sec:Introduction}
\IEEEPARstart{T}{he} routine inclusion of multimodal sensing systems in autonomous vehicles and robotics platform is inevitable. 
Multimodal information improves the reliability and accuracy of many aspects of robotics such as robotic perception, autonomous navigation, dense mapping and localisation as a result of the complementary characteristics of each sensing modality \cite{moghadam2013a,park2017c,park2019, moghadam2008improving}. 
Two of the most important complementary sensing modalities on robotics platforms are LiDARs and visual cameras. A LiDAR provides three-dimensional geometric information while a visual camera represents two-dimensional appearance information.

In order to effectively integrate information (spatially and temporally) obtained from multiple sensing modalities, it is essential to represent them in a common reference frame. Spatial alignment (\ie{} extrinsic calibration) is the process of estimating the relative six degree of freedom (6DoF) transformation (\ie{} rotation and translation) between different sensor coordinate systems.
Time synchronization is the process of estimating any time offset and delay between the different sensing systems.  

The problem of estimation of the rigid body transformation
between the multimodal sensory information (camera and LiDAR) has been extensively studied in the past two decades \cite{kang2019automatic, moghadam2013a, vidas20133d,jinyong2019}. Despite recent developments, fusing multimodal sensory information is still a challenging problem \cite{rehder2014spatio}. 
Previously proposed solutions can provide a reasonable estimation of these parameters but they are limited due to their offline processing requirement \cite{sim2016} and a dedicated artificial calibration target present in a controlled environment~\cite{ahmadyousef2017}. 

In real-world robotic applications, the LiDAR and the camera can be separately located on moving parts of vehicles (semi-rigid), with little or no overlapping field of view and the extrinsic parameters could continuously change, requiring the ability for running calibration routines on-the-fly. As a result, the target-based method cannot be utilized. 
There are methods that do not require a calibration target \cite{rehder2014spatio, moghadam2013a}, however, these methods assume special geometry of the scene (\eg{} lines, planes and points) which are generally only applicable to controlled environments (\ie{} indoor). Furthermore, time synchronization is often ignored in many prior works. The effect of a small time difference resulting either from time lag or time delay is considerably huge in hand-held 3D sensor payloads where the angular velocity is usually very high \cite{ park2017b}. 

To overcome these problems, we propose a targetless, structureless method for LiDAR and visible camera spatiotemporal calibration. Our proposed method has no assumptions about the sensor configurations or parameter initial guesses.  
The proposed method is composed of two stages. The first stage provides a rough closed-form solution for the extrinsic parameters of the unknown system as well as a rough time lag estimation. 
During the second stage, the spatiotemporal parameters are refined by the proposed structureless continuous-time structure-from-motion model. {Due to the structureless approach where the 3D structures are purely constructed from images, there is no requirement to have an overlapping Field of View (FOV) between sensors. }
We demonstrate that the proposed method is capable of acquiring an accurate estimate of the spatiotemporal parameters on different platforms with multiple sensor payload configurations.

\section{Related Work}

The extrinsic calibration between LiDAR and visible cameras has long been studied in robotics. The calibration methods can be grouped into two major categories: target based and targetless.
Target based methods utilize a dedicated, artificial calibration target to be observed simultaneously from LiDAR and visible cameras. 
Zhang~\cite{zhang2004} firstly utilized a planner checkerboard for the extrinsic calibration where normal directions are extracted from a LiDAR point cloud and its corresponding 3D pose is extracted from the 2D image of the checkerboard. 
Three pairs of normal directions and checkerboard poses are enough to calculate the extrinsic parameters. 
By identifying multiple co-planner 3D positions from 3D pointclouds, Guindel \etal{}~\cite{guindel2017} reduced the minimum number of data required for calibration. 
Many other variations of target based methods  \cite{park2014,sim2016} have been proposed to address the limitations of the Zhang~\cite{zhang2004} method, however, due to their requirement of dedicated artificial target objects for calibration, they are not suitable for in-situ calibration where such a calibration target is not available.   

Targetless approaches use natural geometrical features around the scanning area to estimate the extrinsic parameters.
Moghadam \etal{}~\cite{moghadam2013a} utilized the projected 3D line segments on a 2D image in a typical indoor environment. Their method extracts 3D lines from a point cloud and finds matched 2D line segments on the image. Then, matched pairs of line segments are utilized in the pose optimization.
Rehder \etal{}~ \cite{rehder2014spatio} utilized a RANSAC based planer object extraction method. Their method extracts three and more planes from the point cloud and utilizes them for the calibration. However, both methods are limited to a place where such geometrical features are available (\ie{} indoor environment). 

Many of the targetless approaches utilize mutual information to determine the rigid transformation between the coordinate frames. 
Napier \etal{}~\cite{napier2013} applied a mutual information based registration technique to the camera-LiDAR calibration problem. Their method finds the extrinsic parameters that maximizes the mutual information between the intensity edge image of the projected LiDAR point cloud and the image edge pattern. 
Similarly, Gaurav \etal{}~\cite{pandey2015} and Zachary \etal{}~\cite{taylor2012} use mutual information based on raw intensity values and the projected normal point cloud image. 
More recently, Miled \etal{}~\cite{mourad2016} proposed a joint 2D histogram based mutual information calculation.
However, mutual information based approaches are computationally expensive and are not suitable for a temporal calibration. {Furthermore, the approach in this category is only available when enough of a view overlap exists between the camera and LiDAR.}

Temporal calibration is also an important part of multimodal sensor fusion but is often not addressed in previous literature for the reason that many of the systems are either aiming for static scanning \cite{taylor2012} or short-term operation \cite{pavel2018}. 
However, in many practical applications in robotics, the system continuously receives asynchronous estimations from LiDAR and a not connected or unsynchronised visible camera while in motion \cite{guindel2017}. %
This is especially the case for hand-held systems where the angular velocity is rapidly changing. Even a small time lag can result in a considerable misalignment between multimodal sensing systems. 

Most of the extrinsic calibration approaches above, except \cite{rehder2014spatio,moghadam2013a}, ignores the temporal alignment. Temporal calibration is relatively more complicated than extrinsic calibration as asynchronous measurements should be considered. Redher \etal{}~\cite{rehder2014spatio} addressed the problem by introducing a continuous-time trajectory \cite{bosse2009}. However, their method is limited to an initial calibration or batch offline processing only. Some of the recent research \cite{sommer2017} proposed to use a dedicated hardware for the synchronization but a hardware system is not cost efficient and is platform dependent.

\section{Overview}

\begin{figure}[t]
\centering{
\includegraphics[height=.15\textheight]{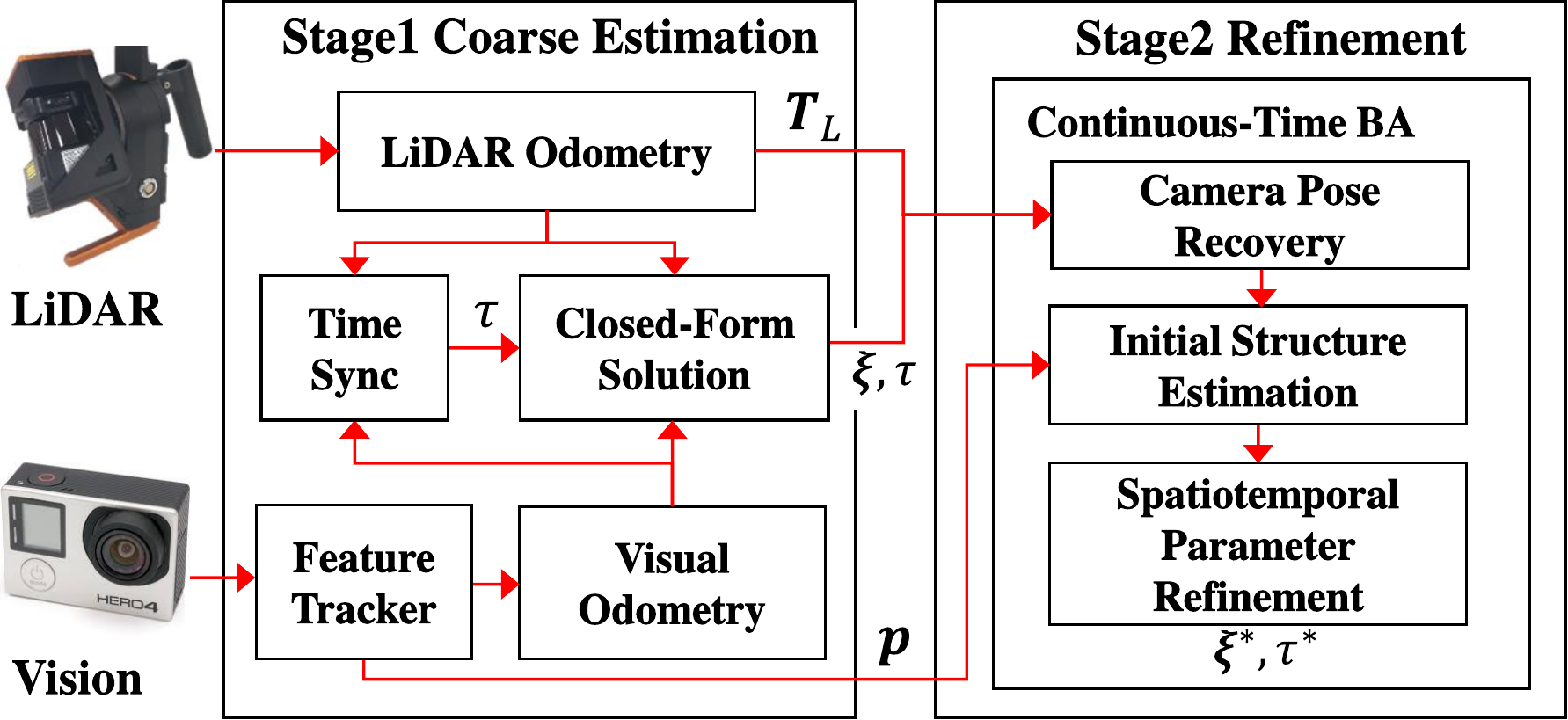}
}
\caption{
Block Diagram of the proposed spatiotemporal camera-LiDAR calibration system.
}
\label{fig:BD}     
\end{figure}

In this section we overview our spatiotemproal calibration method for LiDAR and camera sensory systems as depicted in Fig. \ref{fig:BD}. 
In the first stage of the proposed method, the closed-form solution of the extrinsic is found by the odometry of each sensor. The time stamps of each trajectory are roughly synchronized by detecting a moment where the sensory set starts to move.
We assume that the trajectory of each sensor is estimated independently from the observation of each sensor \eg{} LiDAR trajectory from point cloud registration \cite{park2017c} and camera trajectory from visual odometry by images \cite{mur-artal2015}.
In the second stage, the extrinsic and time lag parameters are refined by reducing the 3D-2D projection error where {the 3D locations of each 2D features are calculated from triangulation rather than projecting 3D LiDAR points onto 2D images or vice versa.}

\section{Proposed Method} 

In this section, we describe our coarse-to-fine approach for spatiotemporal camera-LiDAR calibration.

\subsection{Continuous-Time Trajectory Representation}

Before presenting the details of our calibration method, we briefly explain the continuous-time trajectory representation which models a pose $\textbf{T} \in {SE(3)}$ as a function of time $\tau \in {\mathbb{R}}$,
\begin{equation}
\textbf{T}(\tau):=\begin{bmatrix}
 \textbf{R}(\tau) & \textbf{t}(\tau) \\ 
\textbf{0} &1 
\end{bmatrix} \enspace ,
\end{equation}
where $\textbf{R} \in SO(3)$ and $\textbf{t} \in \mathbb{R}^3$ denote the rotation matrix and translation vector, respectively.
The simplest way to model a continuous-time trajectory is linear interpolation on the manifold. Given two discrete poses $\textbf{T}_i, \textbf{T}_j$ at time $\tau_i, \tau_j$, the pose at time $\tau \in [\tau_i, \tau_j]$ can be interpolated as,
\begin{equation}
\textbf{T}(\tau)= \textbf{T}_i \; e^{\alpha [{\boldsymbol{\xi}}]_\times} \enspace ,
\end{equation}
where the relative pose $[{\boldsymbol{\xi}}]_\times = log \left(\textbf{T}_i^{-1}\textbf{T}_j\right)$ and the interpolation ratio $\alpha = ({\tau-{\tau}_i})/({\tau_j-{\tau}_i})$.

Throughout the rest of this paper, we denote a continuous-time pose as ${}^R \textbf{T}_B(\tau_i)$ and a discrete-time pose as ${}^R \textbf{T}_{B_i}$, where the superscript and subscript represent their reference and body coordinate systems respectively.

\begin{figure}[t]
\centering{
\def\svgwidth{70mm}
\input{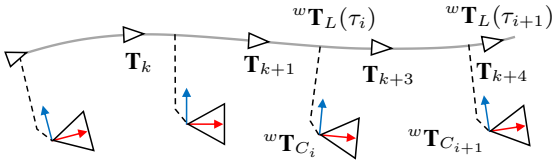}
}
\caption{A continuous-time trajectory representation for a LiDAR and a camera system. The camera poses ${}^w \textbf{T}_{C_i}$ are specified at discrete times $\tau_i$, while the LiDAR poses ${}^w \textbf{T}_{L}(\tau_i)$ are modeled with a continuous-time trajectory $\textbf{T}_k$. The reference world frame, LiDAR frame and camera frame are represented as $w$, $L$, and $C$ respectively.}      
\label{fig:cttraj}       
\end{figure}

{
\subsection{Coarse camera-LiDAR Extrinsic Parameter Estimation}
}

The proposed method utilizes the motion of each sensor to find the rough extrinsic parameters. Thus, we assume that the motion of each sensor is estimated by LiDAR odometry and visual odometry. Then, from each trajectory we extract a set of relative transformations and utilize them for the extrinsic calibration.     

Given a pair of relative transformations in the LiDAR coordinate frame and a pair of relative transformations in the camera coordinate frame, we show that it is possible to calculate the LiDAR to camera transformation with a {closed-form} solution. {In this coarse stage}, we assume that the two trajectories are roughly synchronized, which we will further discuss in the experiment section.

{Fig. \ref{fig:cttraj} depicts a continuous-time trajectory for a system with a LiDAR and a camera. Note that a LiDAR sensor provides continuous observations, while a global-shutter camera produces discrete ones. Thus, the system trajectory is modeled with a continuous-time trajectory in the LiDAR coordinates and parameterized with discrete poses $\textbf{T}_k$. }

Suppose that two camera poses up to scale ${}^{w'}\textbf{T}_{C_i}, {}^{w'}\textbf{T}_{C_j}$ at times $\tau_i, \tau_j$ are given. Then, the relative camera pose is,

\begin{equation}
\label{eq:relative_camera}
{}^{C_i} \textbf{T}_{C_j} =\left({}^{w'}\textbf{T}_{C_i}\right)^{-1} {}^{w'}\textbf{T}_{C_j}
:=\begin{bmatrix}
\textbf{R}_{C_i}& \lambda \textbf{t}_{C_i} \\ 
\textbf{0} & 1 
\end{bmatrix} \enspace,
\end{equation}
where the scale factor $\lambda > 0$ is introduced for monocular camera scale ambiguity.
On the other hand, let us query the corresponding LiDAR poses ${}^w \textbf{T}_{L}(\tau_i), {}^w \textbf{T}_{L}(\tau_j)$ from the estimated continuous-time trajectory. Note that the LiDAR trajectory reference frame $w$ and the camera reference frame $w'$ is not identical. Then, the relative LiDAR pose is,

\begin{equation}
\label{eq:relative_lidar}
{}^{L_i} \textbf{T}_{L_j} =\left({}^{w}\textbf{T}_L (\tau_i)\right)^{-1} {}^{w}\textbf{T}_L (\tau_j)
:=\begin{bmatrix}
\textbf{R}_{L_i}& \textbf{t}_{L_i} \\ 
\textbf{0} & 1 
\end{bmatrix} \enspace.
\end{equation}

{
Assuming that the camera is mounted on the same platform as the LiDAR system, the relative poses are constrained as, 
\begin{equation}
{}^{L_i} \textbf{T}_{L_j} {}^{L} \textbf{T}_C = {}^{L} \textbf{T}_C {}^{C_i} \textbf{T}_{C_j} \enspace,
\label{AXXB} 
\end{equation}
where the camera-LiDAR extrinsic parameters to estimate are,
\begin{equation}
\label{eq:extrinsics}
{}^{L} \textbf{T}_{C}
:=\begin{bmatrix}
 \textbf{R}& \textbf{t} \\ 
\textbf{0} & 1 
\end{bmatrix} \enspace .
\end{equation}
}

Note that the $\textbf{A}\textbf{X}=\textbf{X}\textbf{B}$ in Eq. (\ref{AXXB}) is often noted as the hand-eye calibration \cite{ma2018}. 

{Substituting Eq. (\ref{eq:relative_camera}), (\ref{eq:relative_lidar}) and (\ref{eq:extrinsics}) into Eq. (\ref{AXXB}) gives,
\begin{gather}
\label{eq:rotclosed}
\textbf{R}_L \textbf{R}=\textbf{R} \textbf{R}_C \enspace, \\
(\textbf{I}- \textbf{R}_L)\textbf{t}+\lambda\textbf{R}\textbf{t}_C=\textbf{t}_L \enspace.
\end{gather}
} 
{
The solution to Eq. (\ref{eq:rotclosed}) can be found by aligning correspondences in the manifold. Let $\textbf{R}_{L_k}=e^{[{\textbf{r}}_{L_k}]_\times}, \textbf{R}_{C_k}=e^{[{\textbf{r}}_{C_k}]_\times}$ be the conversion from a rotation matrix to a canonical representation for $k$ pairs of camera-LiDAR relative poses. Then, its covariance is defined as,   
\begin{equation}
\textbf{M} = \sum_{i}^{k} \textbf{r}_{L_i} \: \textbf{r}_{C_i}^\top \enspace .
\end{equation}
}
By Singular Value Decomposition, the {rotation matrix} can be found by decomposing the covariance matrix as,
\begin{equation}
\textbf{R} = (\textbf{M}^\top\textbf{M})^{-1/2}\:\textbf{M}^\top.
\end{equation}

Given the rotation, the translation and scale factor can be found by solving the following linear equation,

\begin{equation}
\begin{bmatrix}
 (\textbf{I}- {\textbf{R}_{L_1}})& \textbf{R} \textbf{t}_{C_1}\\
 \vdots & \vdots\\
 (\textbf{I}- {\textbf{R}_{L_k}})& \textbf{R} \textbf{t}_{C_k}
\end{bmatrix}
\begin{bmatrix}
 \textbf{t} \\ 
\lambda
\end{bmatrix}
=
\begin{bmatrix}
\textbf{t}_{L_1}\\
\vdots \\
\textbf{t}_{L_k}
\end{bmatrix} \enspace .
\label{eq:trans}
\end{equation}

Note, that at least three relative pose pairs {($k > 3$)} are required for the uniquely defined solution. Also, making an adequate amount of roll, pitch, yaw motion is important to estimate the parameters with a sufficient accuracy. The effect of motion will be investigated in Section \ref{sec:experiments}.

\vspace{1mm}
\subsection{Refined Extrinsic and Time Lag Estimation}

\begin{figure}[t]
\centering{
\def\svgwidth{70mm}
\input{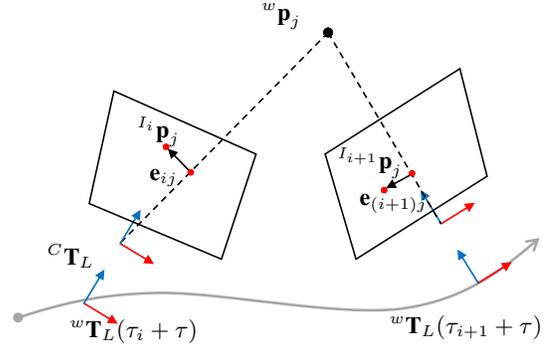}
}
\caption{
{The projection errors $\textbf{e}_{ij}$ of the $j$-th 3D visual feature ${}^w \textbf{p}_j$ onto the $i$-th image frame are used to find the camera-LiDAR extrinsic parameters $\boldsymbol{\xi}$ and time lag $\tau$.}
}
\label{fig:BA}       
\end{figure}

\newcommand \figheight {.135}

\begin{figure*}[t]
\centering{
\includegraphics[height=\figheight\textheight]{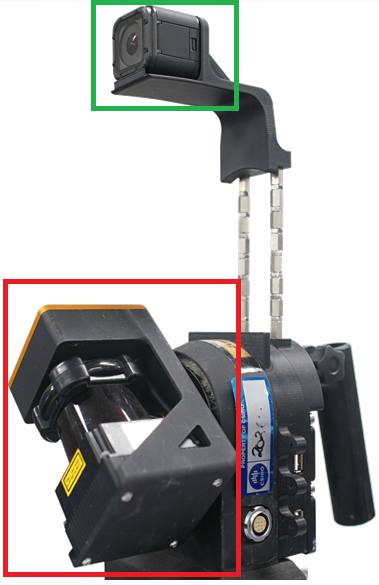}\quad
\includegraphics[height=\figheight\textheight]{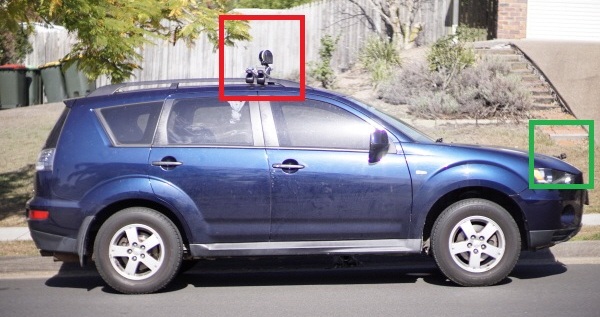}\quad
\includegraphics[height=\figheight\textheight]{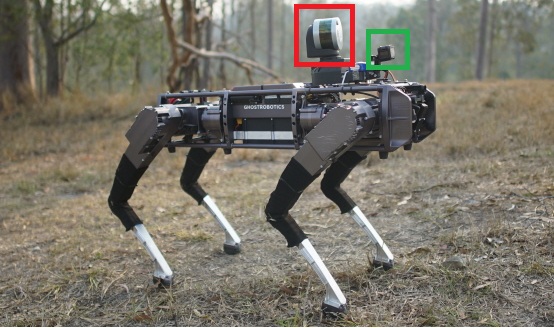}
}
\caption{Sensing Configurations: 
[left]: Hand-held, [middle]: Vehicle, [right]: legged robot used for the spatiotemporal calibration. Two common LiDAR sensors (highlighted in red) are used (UTM-30LX, VLP-16). The added camera (highlighted in green) is an independent camera system and does not share a common clock with the LiDAR.
}
\label{fig:sensorconfig}       
\end{figure*}

The close-form, coarse extrinsic estimation in the previous section reduces an algebraic distance rather than {the geometric} error. 
Thus, once the rough extrinsic {parameters} are {obtained} {as a} closed-form solution, we refine it jointly with the time lag estimation {using non-linear optimization}.

Fig.~\ref{fig:BA} describes the projection errors of visual features. Suppose that the 3D position of a visual feature ${}^w \textbf{p}_j$ {is initialized by triangulation of 2D image features.} Then, the error between the $j$-th feature location on the $i$-th image frame ${}^{I_i} \textbf{p}_j$ and its projection onto the $i$-th image frame is,
\begin{equation}
\textbf{e}{_{ij}}={}^{I_i} \textbf{p}_j-\pi\left({}^{C}\textbf{T}_L {}^{L}\textbf{T}_W (\tau_i+\tau) {}^w\textbf{p}_j\right) \;\;,
\label{eq:proj} 
\end{equation}
where ${}^{C}\textbf{T}_L := e^{[\boldsymbol{\xi}]_\times}$ denotes the extrinsics, $\tau$ represents the  time lag of the LiDAR sensor, and $\pi(\cdot)$ stands for the camera projection.

{We find the optimal parameters $\textbf{x} = \left(\boldsymbol{\xi}, \tau, {}^w\textbf{p}\right)$ which minimizes the following objective function,
\begin{equation}
\textbf{f}(\textbf{x})={\frac{1}{2}\sum_{i}\sum_{j}{\textbf{e}{_{ij}}^\top\boldsymbol{\Sigma}_{ij}^{-1}\textbf{e}{_{ij}}}} \enspace ,
\end{equation}
where $\boldsymbol{\Sigma}_{ij}^{-1}$ denotes a weight matrix of M-estimators.
}

To find the optimal solution iteratively, we apply the Gauss-Newton algorithm with its approximated hessian and gradient,
\begin{equation}
\textbf{H}=\textbf{J}^\top \boldsymbol{\Sigma}^{-1}\textbf{J}, \quad \textbf{g}=\textbf{J}^\top \boldsymbol{\Sigma}^{-1}\textbf{b} \enspace,
\end{equation}
where $\textbf{J} =\frac{\partial \textbf{f}}{\partial \textbf{x}}$ is the Jacobian of the objective function, $\boldsymbol{\Sigma}^{-1}$ is the concatenated weights, and $\textbf{b}$ is the error of the current iteration. Then, the increment is computed as, 
\begin{equation}
\delta\textbf{x}=-\textbf{H}^{-1}\textbf{g},
\label{eq:LM}
\end{equation}
and is iteratively added to the current state as $\textbf{x} \leftarrow \textbf{x} \oplus \delta\textbf{x}$ on the manifold.

{
\subsection{Structureless Optimization Update}
}

For the efficiency of estimation, we introduce the stuctureless approach, where we marginalize the 3D points ${^w\textbf{p}_j}$ from the state estimation. 
First, we divide the state to be estimated as the essential part and the part to be marginalized,
\begin{equation}
\delta\textbf{x}=\begin{bmatrix}
\delta\textbf{x}_c \\ 
\delta\textbf{x}_s
\end{bmatrix}, \;\;
\delta\textbf{x}_c=\begin{bmatrix}
 \delta\boldsymbol{\xi} \\ 
\delta\tau
\end{bmatrix}, \;\;
\delta\textbf{x}_s=\begin{bmatrix}
 {\delta{}^w\textbf{p}_{ 1}}  \\ 
  ... \\ 
{\delta{}^w\textbf{p}_{ J}} 
\end{bmatrix},
\end{equation}
where $\delta\textbf{x}_c$  are the extrinsics and time lag and $\delta\textbf{x}_s$ are the set of 3D points to be marginalized. Then, we can decompose the Eq. (\ref{eq:LM}) into two parts as,
\begin{equation}
\begin{bmatrix}
 \textbf{H}_{cc}& \textbf{H}_{cs} \\ 
\textbf{H}_{sc} & \textbf{H}_{ss} 
\end{bmatrix}
\begin{bmatrix}
 \delta\textbf{x}_c \\ 
\delta\textbf{x}_s  
\end{bmatrix}= -
\begin{bmatrix}
 \textbf{g}_{c} \\ 
\textbf{g}_{s}  
\end{bmatrix}.
\end{equation}

Then, we apply the Schur complement to marginalize the subset of variables,
\begin{gather}
\bar{\textbf{H}}_{cc}=\textbf{H}_{cc}-\textbf{H}_{cs}{\textbf{H}_{ss}}^{-1}\textbf{H}_{sc} \enspace,\\
\bar{\textbf{g}}_{c}=\textbf{g}_{c}-\textbf{H}_{cs}{\textbf{H}_{ss}}^{-1}\textbf{g}_{s}\enspace.
\end{gather}

Finally, the marginalized increment is given by solving the following linear equation,
\begin{equation}
\bar{\textbf{H}}_{cc}\:\delta\textbf{x}_c= -\bar{\textbf{g}}_{c}.
\end{equation}
The structure part should be recalculated for each iteration \cite{triggs1999}.
It is also possible to estimate the time lag only by further marginalizing the extrinsic parameters.

\begin{figure}[t]
\centering{

\includegraphics[width=.47\textwidth]{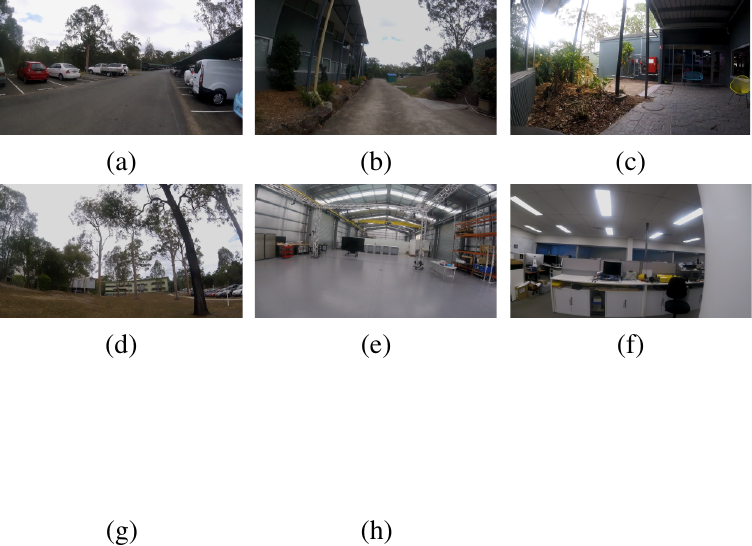}
}
\caption{Screenshots of the datasets in Table \ref{tbl:datasets}. (a) to (d): different environments, (e): no view overlap, (f) to (g): different platforms and baseline lengths.  }
\label{fig:datasetvis}       
\end{figure}

\begin{figure}[t]
\centering{
\def\svgwidth{70mm}
\includegraphics[height=.13\textheight]{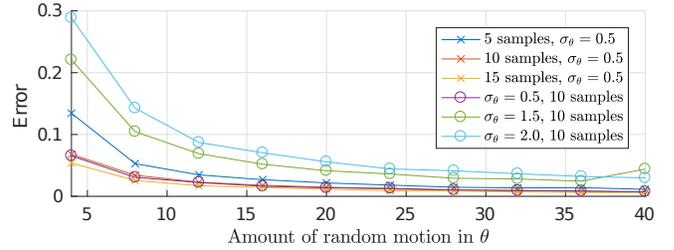}
}
\caption{
Rotation error of the closed-form solution with various simulated motions. The higher motions in roll, pitch and yaw directions produce the more accurate rotation estimations for various sample and noise cases. The unit of y-axis is the angular vector norm.
}
\label{fig:angvserr}       
\end{figure}

\vspace{2mm}
\section{Experiments}
\label{sec:experiments}

In this section, we present quantitative and qualitative performance analysis of the proposed method with both simulated and real data in different sensor configurations. In particular, we compare the accuracy of the closed-form solution in the coarse stage with the optimized solution in the refinement stage. Comparison of the proposed method on different types of environment are presented.  

\subsection{Experiment Setup}

Fig.~\ref{fig:sensorconfig} illustrates multiple camera-LiDAR sensor configurations mounted on hand-held, vehicle and legged robot used for our experiments. We have used two common LiDAR sensors, highlighted in red, (UTM-30LX, VLP-16). The added camera (highlighted in green) is an independent camera system and does not share a common clock with the LiDAR. Video recording is manually triggered by an operator which yields an unknown initial time lag. 

For the real data experiment, we recorded multiple datasets (Fig. \ref{fig:datasetvis}) in different environments and with different sensor payload configurations. 
The dataset (a) to (d) were recorded in multiple environments with the same hand-held single beam LiDAR setup (Fig.~\ref{fig:sensorconfig}  left). In the dataset (e), the camera and LiDAR were pointing in the opposite direction which is presented to demonstrate the ability of the calibration where the camera and LiDAR do not share a view at all.
For the demonstration of the proposed method on various robotics platforms with different baseline lengths, the dataset (f) to (h) were collected with a handheld device with a longer-baseline (f), vehicle (g), and a legged robot(h) as shown in Fig.~\ref{fig:sensorconfig}  middle and right. The multi-beam LiDAR is utilized for the dataset (f) to (h).

\begin{table*}[t]
\caption{Calibration Accuracy According To Environment}
\begin{center}

\begin{tabular}{lcccccccc}
\hline\noalign{\smallskip}
 Dataset & \multicolumn{1}{c}{(a) Carpark}  & \multicolumn{1}{c}{(b) Outdoor}  & \multicolumn{1}{c}{(c) In/outdoor}   & \multicolumn{1}{c}{(d) Open Space}   & \multicolumn{1}{c}{(e) No Overlap}  & \multicolumn{1}{c}{(f) Long Base} & \multicolumn{1}{c}{(g) Vehicle}    & \multicolumn{1}{c}{(h) Legged Robot}   \\
\hline\noalign{\smallskip}

$e_\textbf r$($e_\textbf t$) coarse	&  7.0(0.021) 	& 10.4(0.018) 	&       106.4(0.050) 	&  38.8(0.057) 	& 120.4(0.055) 	&11.8(0.007)   & 12.4(0.630)& 31.0(0.178) \\
$e_\textbf r$($e_\textbf t$) refine  	& 2.2(0.012) 	& 4.2(0.012) 	&       3.4(0.016) 	&    2.4(0.003) & 2.2(0.009)    & 7.1(0.009) 	&	 4.0(0.05)& 2.8(0.019) \\
$e_d$         				&  1ms 		& 1ms 		& -153ms		&-186ms		& -393ms	&  -15ms 		& -267ms & -183ms  \\
baseline         			&  0.22		&  0.22		&  0.22			&  0.22		& 0.22		&   1.1			&  2.5&  0.20  \\

\noalign{\smallskip}\hline\noalign{\smallskip}   

\end{tabular}
\end{center}
\begin{tablenotes}
\small
\item  {\footnotesize 
Comparison of the spatiotemporal calibration error on multiple datasets from different environments. $e_\textbf r$($e_\textbf t$) coarse and $e_\textbf r$($e_\textbf t$) refine represent the extrinsic estimation error respectively for before and after the refinement. $e_d$ represents the difference between initial time lag estimated by the method in Fig. \ref{fig:timesync} and refined time lag. The unit of the baseline length is in meters. $e_\textbf r$ stands for the rotational error in radians with $10^{-3}$ scale, $e_\textbf t$ for the translational error in meters, $e_ \tau$ for the time lag error in milliseconds.}

\end{tablenotes}
\label{tbl:datasets}
\end{table*}

\subsection{Closed-Form Solution in the Coarse Stage}
\subsubsection{Experiments with Simulated Data}
In the first experiment, we evaluate the accuracy of the estimated rotational part of the extrinsic parameters based on how many relative pose samples between camera-LiDAR are needed as well as the amount of variations required to excite relative poses between two time stamps.

The experiment in Fig. \ref{fig:angvserr} shows that the most dominant component is the amount of motion in the relative pose but this did not improve after some point (around 30 degrees). Also, the more number of samples after 15 did not significantly improve the accuracy of the rotation parameter estimation. With the higher noise level (line with circles) the overall rotation estimation is more noisier but showed the similar pattern.
Based on this observation we utilized 10 relative pose samples for the experiment with the real data in the next section. Also, we made the motion larger than 25 degrees.

\subsubsection{Experiments with Real Data}
\begin{figure}[t]
\centering{
\includegraphics[height=.10\textwidth]{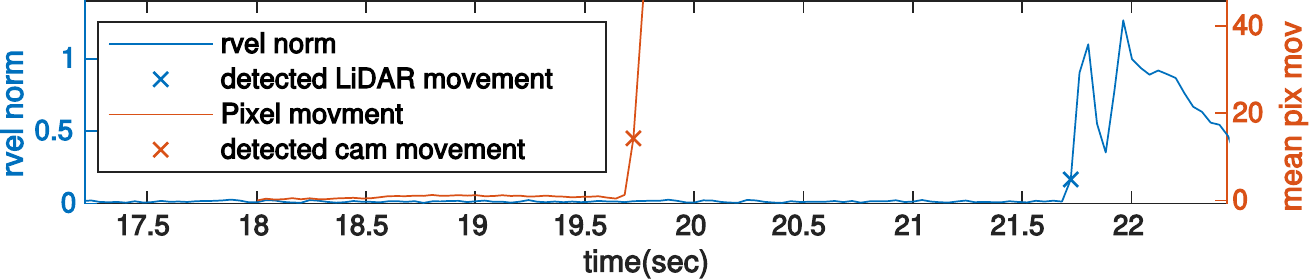}

}
\caption{Example of initial time synchronization by detecting the start of the motion. This rough synchronization is enough for generating a reasonable initial guess on the extrinsic as given in Table \ref{tbl:syncerror}.  
}
\label{fig:timesync}       
\end{figure}

We also performed experiments with real data to evaluate the accuracy of the close-form solution for estimating extrinsic parameters. 
The mono camera trajectories were estimated by visual odometry \cite{mur-artal2015}, while the continuous-time trajectory of the LiDAR sensor were estimated by our previous work \cite{park2017c, park2019}. 
Prior to the closed-form solution calculation, two trajectories are roughly synchronized automatically by detecting the start of the motion using the rotation velocity of the LiDAR trajectory and mean feature movement in the 2D image space as depicted in Fig.~\ref{fig:timesync}. The amount of the time lag estimation error in the coarse stage is given in Table \ref{tbl:datasets} $e_d$ row. 
We assumed that the LiDAR motion distortion is compensated in the odometry estimation \cite{ park2019}. 

The first row of Table \ref{tbl:datasets} summarizes the accuracy in extrinsic parameter estimation with different types of real datasets where each estimation is compared to the ground-truth. The ground-truth extrinsic parameters for quantitative evaluation are obtained by manually selecting 3D points from the LiDAR point cloud and their corresponding 2D points in images. A non-linear least square optimization is used to minimize the projected distance between the 3D and 2D points and to obtain the ground-truth. We validated the ground-truth estimation visually over multiple frames across time. The ground-truth time lag is estimated by the global time lag optimization method in \cite{pavel2018} and is manually verified.
 
Table \ref{tbl:syncerror} shows how the synchronization error affects the accuracy of coarse extrinsic parameter estimation by introducing artificial time lag between two sensing modalities. 
Results suggest that the extrinsic parameter estimation in the coarse stage is less sensitive to relatively high time lag ($\pm$0.5 sec) while being able to estimate good enough initial guesses for the refinement stage. 

\begin{table}[t]
\caption{Accuracy According To The Time Lag}
\begin{center}

\begin{tabular}{lcccccc}
\hline\noalign{\smallskip}
 Time lag & \multicolumn{1}{c}{-0.5s} & \multicolumn{1}{c}{-0.3s} & \multicolumn{1}{c}{-0.1s} & \multicolumn{1}{c}{0.1s} & \multicolumn{1}{c}{0.3s} & \multicolumn{1}{c}{0.5s}                                  \\
\hline\noalign{\smallskip}

$e_\textbf{r}$    &148.5  &108.1&   65.8&      64.0&   89.7&   80.1 \\
$e_\textbf {t}$   &  0.06&    0.04&    0.05&       0.06&    0.04&    0.08 \\
\noalign{\smallskip}\hline\noalign{\smallskip}   

\end{tabular}
\end{center}
\begin{tablenotes}
\small
\item  {\footnotesize 

Effect of the synchronization error in the coarse stage. $e_\textbf{ r}$ stands for the rotational error in radians with $10^{-3}$ scale, $e_\textbf {t}$ for the translational error in meters. 
 }
\end{tablenotes}
\label{tbl:syncerror}

\end{table}

\begin{figure}[t]
\centering{
\def\svgwidth{70mm}
\includegraphics[height=.18\textwidth]{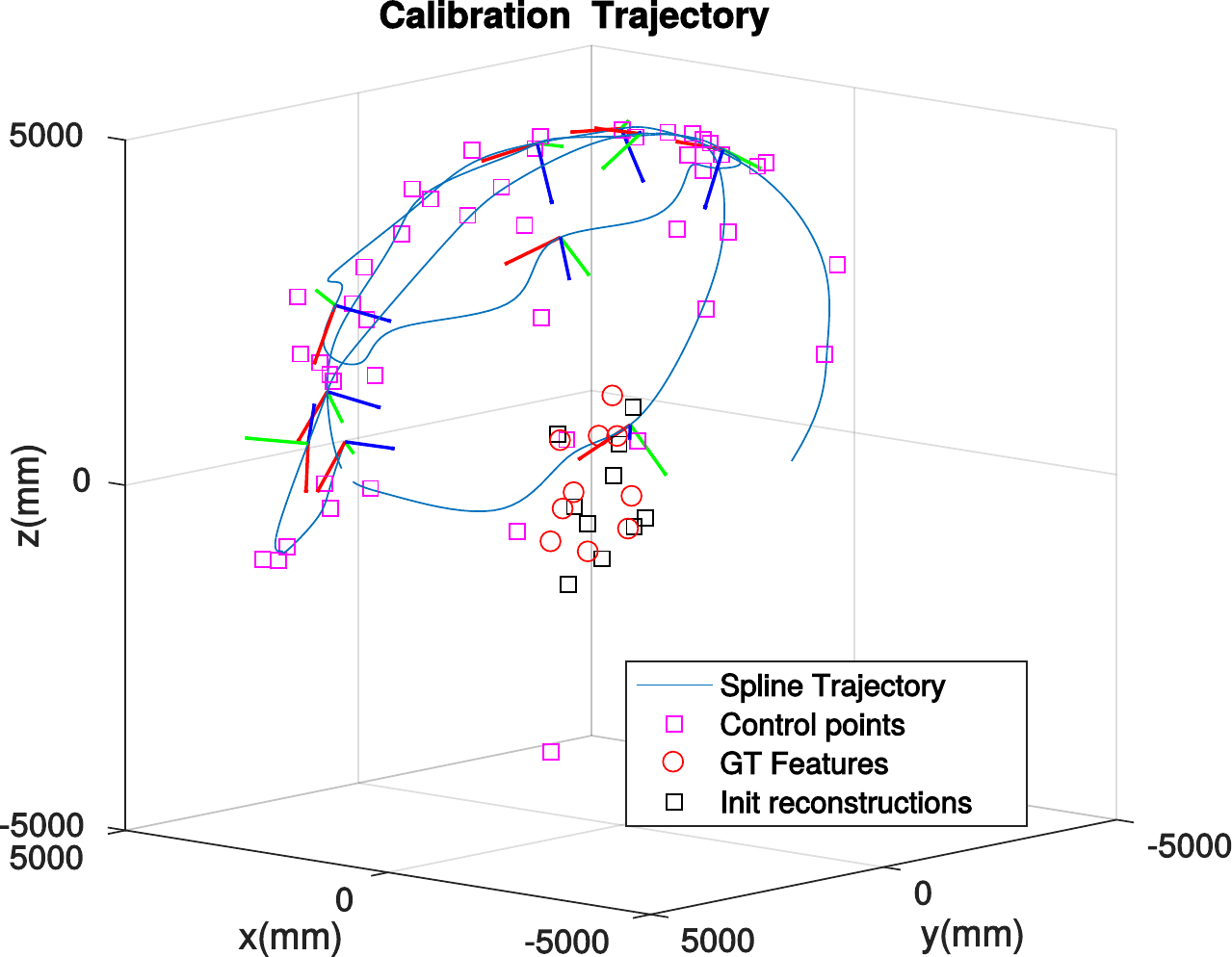}
\includegraphics[height=.18\textwidth]{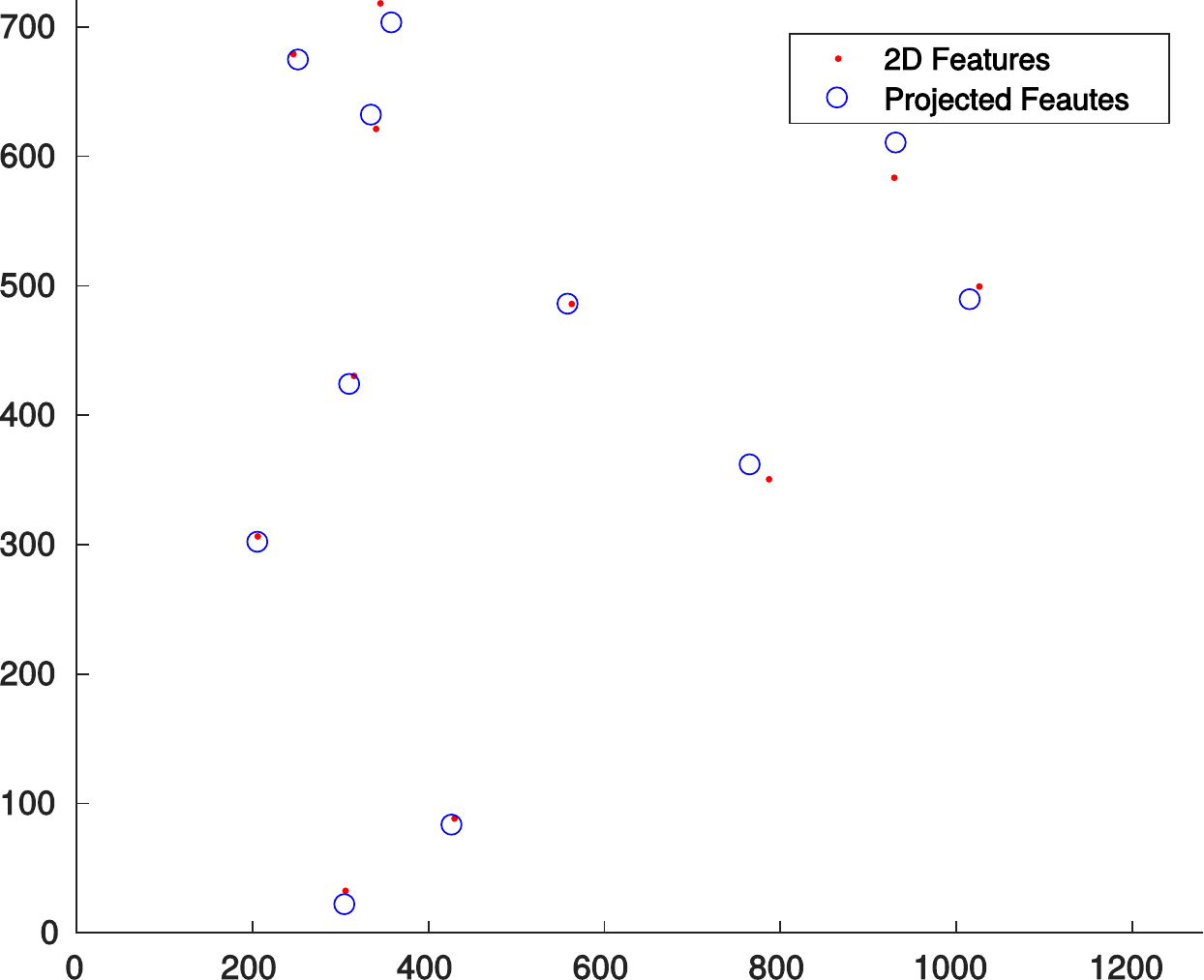}
}
\caption{
[left]: Simulated trajectory with 3D features, [right]: Noise corrupted 2D features in the image space and corresponding 3D projected points after calibration.
}
\label{fig:basim}       
\end{figure}

\begin{table}[t]
\caption{Accuracy According To The Number of Frames Used}
\begin{center}

\begin{tabular}{lrrrr}
\hline\noalign{\smallskip}
 No.Frame  & \multicolumn{1}{c}{10 Frames} & \multicolumn{1}{c}{20 Frames} & \multicolumn{1}{c}{30 Frames} & \multicolumn{1}{c}{50 Frames}  \\
\hline\noalign{\smallskip}
$e_\textbf r$ & 15.4 (16.8) & 7.1 (11.9)& 5.1 (9.50)& 2.0 (5.20) \\
$e_\textbf t$ &  0.2 (0.26) & 0.09 (0.19)& 0.07 (0.20)& 0.01 (0.08) \\
$e_\tau$          &  3.5 (4.70) & 1.4 (3.20)& 1.0 (2.60)& 0.4 (2.00) \\
\noalign{\smallskip}\hline\noalign{\smallskip}   
\end{tabular}
\end{center}

\begin{tablenotes}
\small
\item  {\footnotesize 

The errors (means and variances) in optimally estimated extrinsic parameters and time lag with the simulated data in Fig.  \ref{fig:basim}. The errors decrease as more frames are added to the optimization. $e_\textbf r$ stands for the rotational error in radians with $10^{-3}$ scale, $e_\textbf t$ for the translational error in meters,  $e_ \tau$ for the time lag error in milliseconds.
}
\end{tablenotes}
\label{tbl:basim}

\end{table}

\begin{figure*}[t]
\centering{
\includegraphics[]{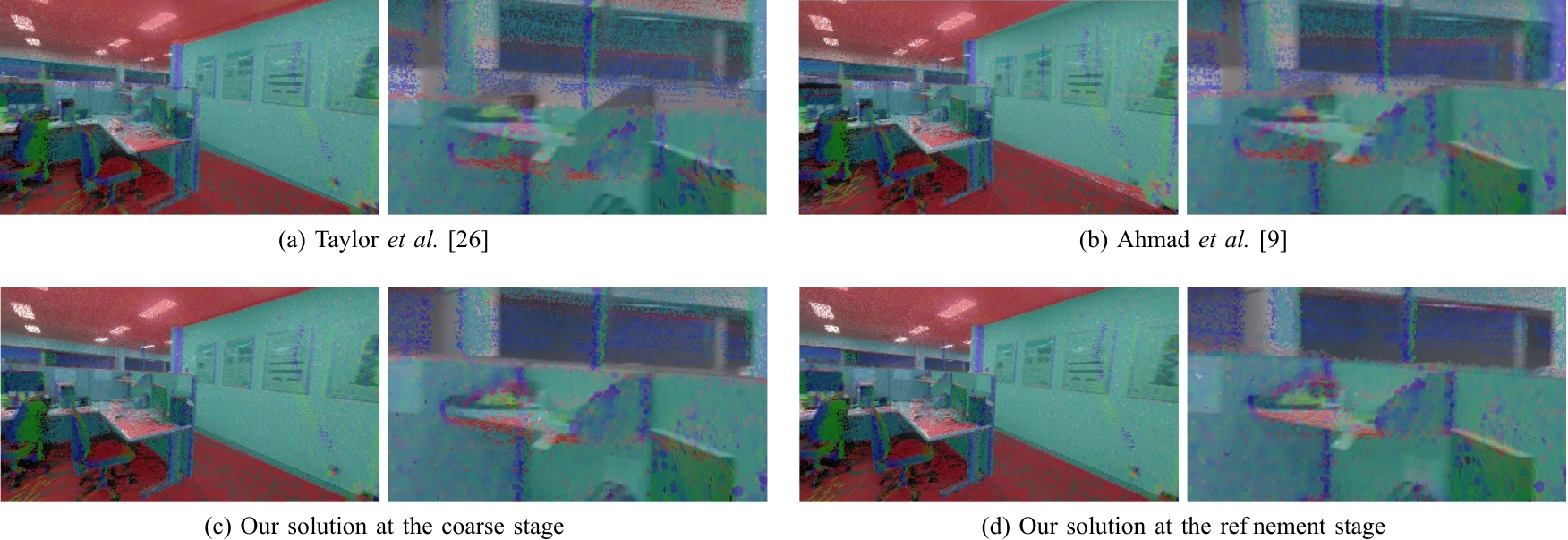}
}
\caption{
Qualitative comparison of accuracy between different methods. For each method, the blend of the captured image and normal map is shown on the left, while the center area is enlarged on the right.
}
\label{fig:viscomp}       
\end{figure*}

\subsection{Extrinsic and Time Lag Estimation in the Refinement Stage}

\subsubsection{Experiments with Simulated Data}

As the exact ground-truth for extrinsic parameters and time lag estimation in real datasets are often unknown, we performed simulated experiments to demonstrate the accuracy of the proposed refinement stage.
As shown in Fig. \ref{fig:basim}, we simulated a camera with a resolution of 1280$\times$720 at 20 Hz.
We mimicked a hand-held motion for the system trajectory and introduced a Gaussian noise of $\mathbf{\Sigma}=diag(5,5)$ pixels to the projected visual feature locations.
The same extrinsic values of the real device are utilized with an extrinsic error that are randomly generated according to the expected error variance of the close-form solution. The time lag is set within 10 milliseconds. We repeated the experiment 50 times with random trajectories and results are summarized in Table \ref{tbl:basim}. The test is repeated with different numbers of frames. Frames are evenly selected over the entire trajectory.    

The experiment shows that with more number of frames included a better accuracy is achieved as well as preventing the chance of having a trajectory that does not have enough coverage of roll, pitch, yaw. But additional frames increases computational complexity. Selection of the frame number should be differentiated according to the angular and linear velocity of the trajectory that will be utilized for the calibration. Based on the simulation result in Table \ref{tbl:basim}, we have utilized 10-30 key frames for the hand-held and robot mounted experiments in the next section.

\subsubsection{Experiments with Real Data}
  
The estimated extrinsic parameters are further refined with the proposed continuous-time structureless bundle adjustment. 
For the proposed iterative optimization method, 2D features are tracked by KLT \cite{tomasi1991} and only 30 distinctive frames are utilized to estimate the spatiotemporal parameters. Outliers are rejected by the M-estimator. To prevent a sub-optimal result due to the initial synchronization error, only the time lag is refined in the beginning by marginalizing the extrinsic and structure from the state. Then, the extrinsic is jointly optimized with the time lag after updating the image time stamps. The extrinsic is updated by the closed-form solution before the joint estimation. The same process of obtaining the ground-truth for extrinsic parameters in the coarse stage is used for the refinement stage. 

The comparison of the extrinsic calibration result between the different state-of-the-art methods is given in Table \ref{tbl:gtcomp}, along with the visual comparison in Fig.~\ref{fig:viscomp}.
For a comparison, closed-form solution by a normal trajectory segment {\cite{taylor2016}}, target-based calibration method {\cite{ahmadyousef2017}}, and refined extrinsic parameters are utilized.
The qualitative comparison in Fig.  \ref{fig:viscomp} shows that the image to point cloud alignment is most accurate when it is refined.

The extrinsic estimation error after the refinement is given in the second row of Table \ref{tbl:datasets}. 
As the proposed method does not require any shared geometric features between the LiDAR and camera, the method works at various types of places ((a),(b) outdoor, (c) indoor, (b),(d) place with lack of geometrical features) and configurations ((e) single beam LiDAR without view overlap). Also, the experiment result with a longer baseline (f) showed a similar performance. 
The result in (g), (h) indicates that the proposed method is suitable for the on-wake-up self-calibration on a legged robot and a vehicle (Fig. \ref{fig:sensorconfig} (a) [middle, right]).

The time lag error after the refinement stage could not be evaluated as the accuracy of the ground-truth estimation method \cite{pavel2018} is in a similar range to that of our refinement stage. This is due to the fact that in the method time lag and extrinsic cannot be jointly estimated. The ground-truth estimation of the extrinsic parameters are not affected by the time lag if a stationary key frame is used whereas the opposite is not true. Instead, we provide a video\footnote{https://youtu.be/ESfP8Uj0v3g} that qualitatively demonstrates the temporal alignment quality.

The time lag calibration of above is only valid within the utilized image frames. Since, the time lag continuously changes, the time lag should be tracked on-the-fly. The result in Table \ref{tbl:bareal_time} demonstrates such ability of the proposed method without the dedicated sensor motion. In this stage, we only run the refinement operation without the first stage. For the experiment, the randomly selected five trajectory segments which were extracted from a normal hand-held motion with three seconds length and 30 key frames are utilized. The motion speeds are 0.7 $rad/s$ and 0.9 $m/s$. The result in Table \ref{tbl:bareal_time} implies that the proposed method was able to estimate the parameters even with relatively large time lags without effecting the extrinsic parameters. 

\begin{figure}[t]
\centering

\includegraphics[height=.20\textheight]{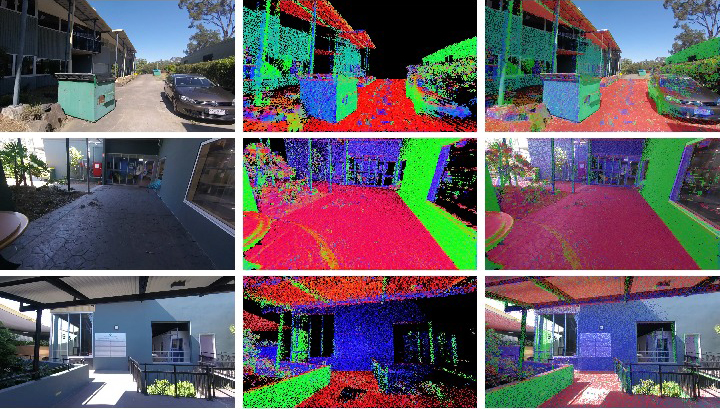}
\caption{
(left): captured images, (middle): surfel clouds projected on the image, (right): blend of the captured images and projected surfel clouds. The color in the surfel clouds represents normal directions.   
}
\label{fig:vis_exampl}       
\end{figure}

Some of the detailed scenes are presented in Fig.  \ref{fig:vis_exampl} using the estimated spatiotemporal parameters to show the camera and LiDAR alignment quality by blend of the captured images and the projected surfel clouds colored by the normal directions.

\begin{table}[t]
\caption{Robustness Against The Time Lag }
\begin{center}
\begin{tabular}{lrrrr}
\hline\noalign{\smallskip}
 Time lag & \multicolumn{1}{c}{33ms} & \multicolumn{1}{c}{66ms} & \multicolumn{1}{c}{99ms} & \multicolumn{1}{c}{133ms}                                  \\
\hline\noalign{\smallskip}
$e_\textbf r$        & 4.48 (1.48) &4.47 (1.45)& 4.46 (1.43)& 4.47 (1.45) \\
$e_\textbf t$        & 0.05 (0.03) &0.04 (0.03)& 0.04 (0.03)& 0.04 (0.03)\\

$e_\tau$   &  -0.03 (0.01) & -0.08 (0.03)& -0.10 (0.04)& -0.15 (0.03) \\
\noalign{\smallskip}\hline\noalign{\smallskip}   

\end{tabular}
\end{center}

\begin{tablenotes}
\small
\item  {\footnotesize 

Robustness against the time lag in estimating spatiotemporal parameters. Utilized camera records in 30 frame per second where frame interval is around 33ms. $e_\textbf r$ stands for the rotational error in radians with $10^{-3}$ scale, $e_\textbf t$ for the translational error in meters,  $e_\tau$ for the time lag error in milliseconds.
}
\end{tablenotes}
\label{tbl:bareal_time}
\end{table}

\begin{table}[t]
\caption{Calibration Accuracy Comparison }
\begin{center}
\begin{tabular}{lccc}
\hline\noalign{\smallskip}
 Methods & Taylor \etal{}  \cite{taylor2016}  &   Ahmad \etal{} \cite{ahmadyousef2017}& Ours                                  \\
\hline\noalign{\smallskip}
$e_\textbf r$ & 45.30 & 80.10    &  \textbf{7.00} \\
$e_\textbf t$   &   0.150   &  0.120 &   \textbf{0.01} \\
\noalign{\smallskip}\hline\noalign{\smallskip}
\end{tabular}
\end{center}
\begin{tablenotes}
\small
\item  {\footnotesize 

Accuracy comparison with different methods. Manually tuned extrinsic parameters and time lag are utilized as ground-truth. $e_\textbf r$ stands for the rotational error in radians with $10^{-3}$ scale, $e_\textbf t$ for the translational error in meters.}
\end{tablenotes}
\label{tbl:gtcomp}
\end{table}

\section{Discussion}

The translation estimation error ratio is relatively higher with the short baseline setups. Given the other experiments with the longer baseline shows the similar error range, it is reasonable to assume that the 0.5 to 4cm error in translation estimation is the estimation accuracy lower bound with our sensor setup and estimation uncertainty in feature tracking and LiDAR trajectory.  

The comparison of the extrinsic estimation in the coarse and refinement stage suggests that conventional motion-based only methods yields relatively high uncertainty in rotation estimation. It is due to the fact that even a small rotational motion is amplified in the refinement stage by being projected on the image and therefore rotation is more observable during the optimization.

The extrinsic estimation of the dataset (a) to (d) has been performed with the identical sensor setup at the different environments. The mean and standard deviation of the rotation and translation error after the refinement is 4.0(1.6)$\times10^{-3}$ radian and 0.017(0.012) meters respectively with the standard deviation in () whereas the coarse stage shows 13.1(12.7)$\times10^{-3}$ radian and 0.019(0.007) meters respectively. 
This implies that the proposed refinement stage is able to provide robust estimation of the extrinsic parameters regardless of the environment.

\subsection{Limitation and Future works}
While the proposed method is flexible, it has a couple of practical {limitations}.

{Time lag estimation:} the time lag parameter is not observable when the platform is stationary{. Time lag uncertainty  grows during the stationary mode.} Thus, for a system with an independent camera, this could be problematic when the platform resumes to move after {a long period of time.}
{Dedicated motion: }
while in {many robotic platforms, such as legged robots, it is easier to move for calibration, there are platforms where it is not easy to do so}. For example, with the vehicle type platform it is not easy to make adequate motion in a short time. Also, the proposed method requires 3D geometrical features and trackable visual features to exist in the scene. The lack of these geometrical features can cause localization slips in LiDAR odometry estimation which affects the accuracy of the calibration result. The lack of trackable visual features presents a similar issue. Thus, observability test and input data analysis that judge if the collected data is enough to carry out the calibration should be further studied for practical application.

\section{Conclusion}
\label{sec:6}

In this paper, we proposed a spatiotemporal calibration method for a camera-LiDAR system. 
The proposed method provides an efficient way for camera-LiDAR calibration without a dedicated calibration target or view alignment. This is achieved by introducing a structureless refinement stage where the key advantage is that 3D points for reprojection are being calculated only from triangulation of 2D features, therefore, the proposed method does not require a known target or view alignment between the LiDAR and camera for matching features. Also, the coarse stage provides a close enough spatiotemporal initial guess which prevents the refinement stage from failure or getting stuck in a local minima. 
Through the experiments, we demonstrated the utility of the method on various robotics platforms and scenarios.
While the proposed method is more flexible than existing solutions, the results illustrate a decent accuracy for the point cloud colorization under situations where trackable and stable visual patterns exist in the view.

\section*{Acknowledgments}
The authors gratefully acknowledge funding of the project by the CSIRO and QUT. Support from several members of the CSIRO Robotics and Autonomous Systems including Tam Benjamin, Gavin Catt and Mark Cox are greatly appreciated.

\balance{}

{\small
        \bibliographystyle{IEEEtran}
        \bibliography{ref}
}

\end{document}